\documentclass[letterpaper]{article}
\usepackage{flairs}
\usepackage{times}
\usepackage{helvet}
\usepackage{courier}
\usepackage{url}
\usepackage{graphicx}
\graphicspath{ {figures/} }
\usepackage{caption}
\usepackage{subcaption}
\usepackage{wrapfig}
\usepackage{multirow}
\usepackage{tabularx}
\usepackage{paralist}
\usepackage{colortbl}
\usepackage{dingbat, pifont}
\definecolor{Gray}{gray}{0.9}

\usepackage{array} 

\begin{document}
%
\title{Enhanced Multimodal Content Moderation\\ of Children's Videos using Audiovisual Fusion}
\author{
\begin{tabular}{llll}
Syed Hammad Ahmed & Muhammad Junaid Khan & Gita Sukthankar\\
\end{tabular}
\\
Department of Computer Science, University of Central Florida, Orlando, FL USA\\
\{syed.hammad.ahmed, muhammad.junaid.khan, gita.sukthankar\}@ucf.edu\\ 
}

\maketitle

\begin{abstract}
Due to the rise in video content creation targeted towards children, there is a need for robust content moderation schemes for video hosting platforms. A video that is visually benign may include audio content that is inappropriate for young children while being impossible to detect with a unimodal content moderation system. Popular video hosting platforms for children such as YouTube Kids still publish videos which contain audio content that is not conducive to a child's healthy behavioral and physical development. A robust classification of malicious videos requires audio representations in addition to video features. However, recent content moderation approaches rarely employ multimodal architectures that explicitly consider non-speech audio cues. To address this, we present an efficient adaptation of CLIP (Contrastive Language–Image Pre-training) that can leverage contextual audio cues for enhanced content moderation. We incorporate 1) the audio modality and 2) prompt learning, while keeping the backbone modules of each modality frozen. We conduct our experiments on a multimodal version of the MOB (Malicious or Benign) dataset in supervised and few-shot settings. 
 
\end{abstract}

\section{Introduction}

There has been a remarkable surge in both the creation and consumption of video content on the internet, making automated content moderation indispensable for video sharing platforms. More than 80\% of the internet usage globally is from video data streaming and downloading \cite{Shewale_2024a}.  In many countries, content moderation is mandatory for video hosting platforms to remain compliant with legal regulations. 
Recent findings indicate that around 2,500 videos are published per minute on YouTube, which is the largest video sharing platform \cite{ytusagedaily}.  Unfortunately, there is a significant amount of content which can be detrimental to emotional and psychological well-being of the viewer. 
 Extensive consumption of malicious video content has socio-economic impacts in addition to its effects on individual viewers. 

\begin{figure}[htp]
    \centering
    \includegraphics[width=0.85\columnwidth]{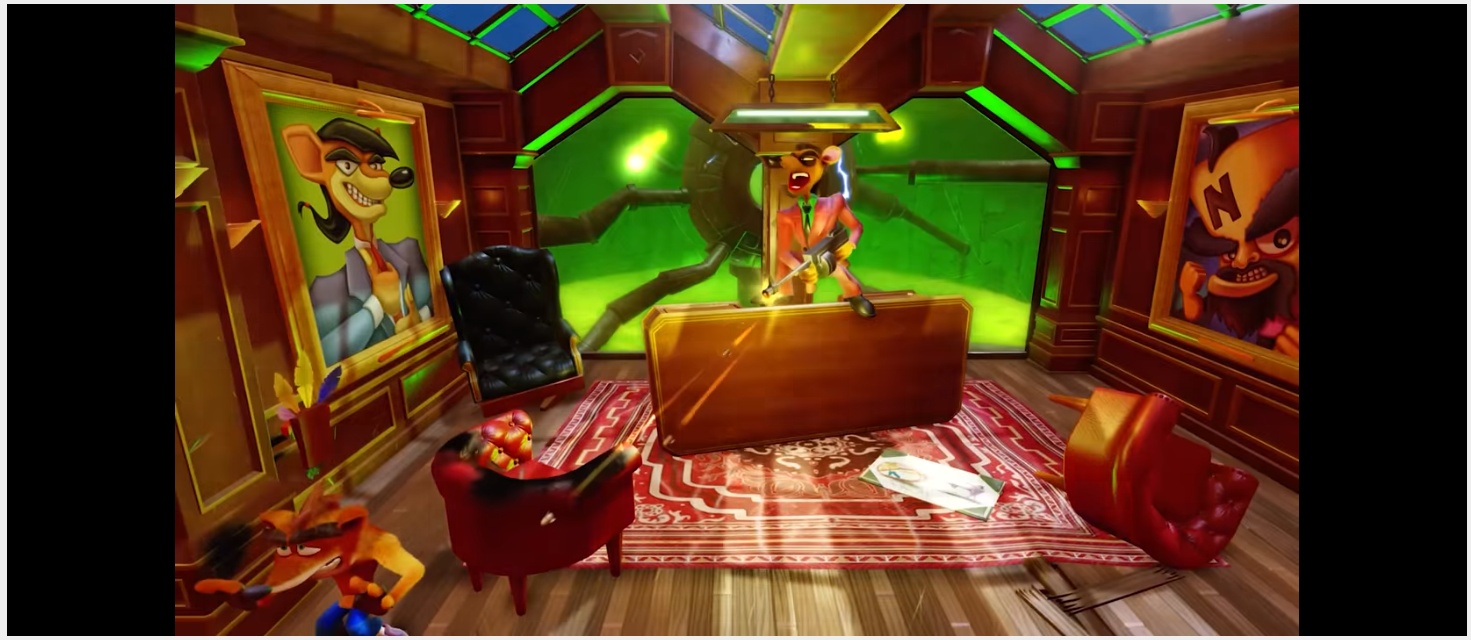}
    \caption{One of the malicious video examples currently available on the YouTube Kids platform that shows a furious cartoon character shooting at other cartoon characters with a machine gun.}
    \label{fig:ytkmachinegun}
\end{figure}

\begin{figure}[htp]
    \centering
    \includegraphics[width=0.85\columnwidth]{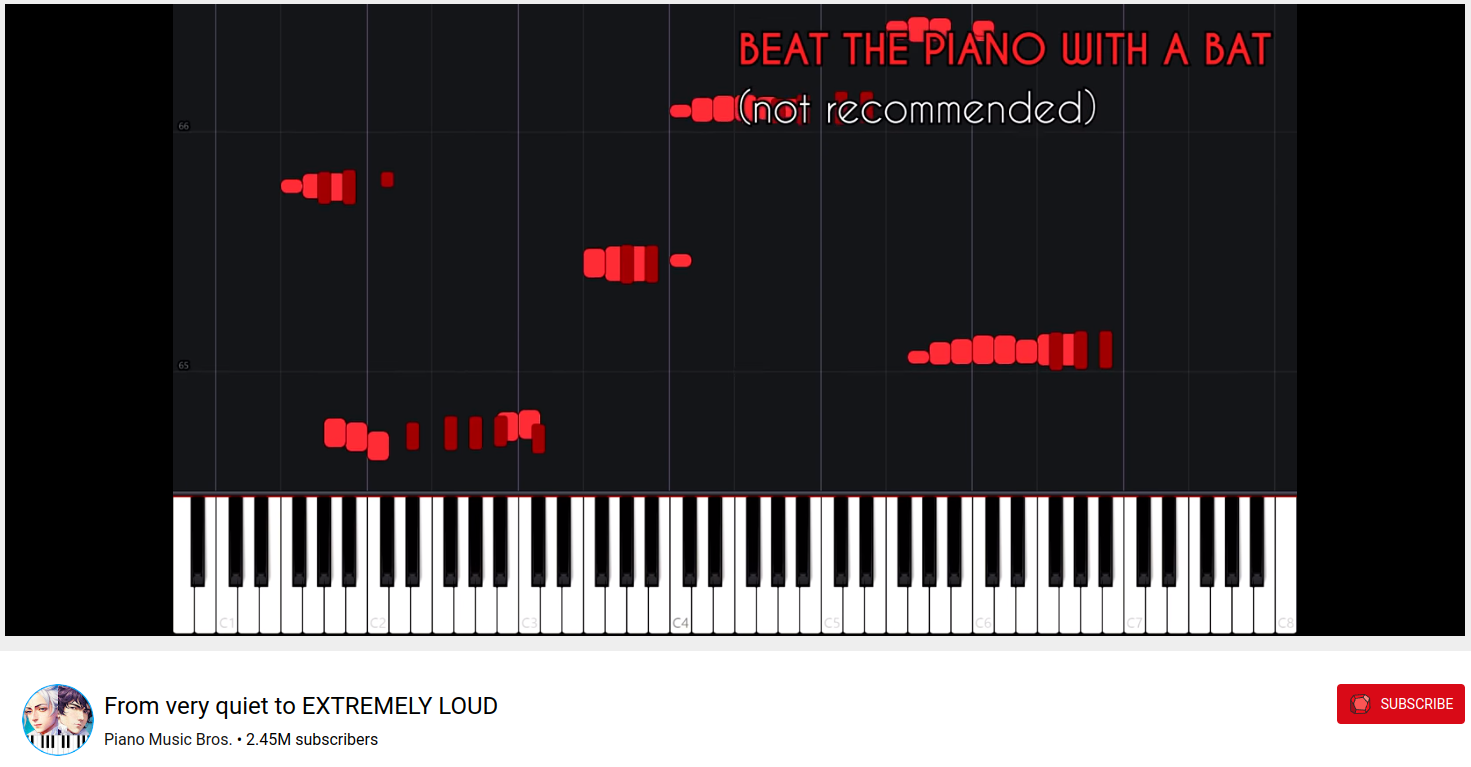}
    \caption{A sample video which includes fast and loud piano notes. Usually in such videos, there is high tempo, a lack of rhythm and varying pitches. The video also includes bright and striking hues, and suggests the violent action of ``hitting the piano with a bat''.}
    \label{fig:ytkpiano}
\end{figure}

Lawsuits have been filed against various online content sharing platforms for intentionally fostering addiction among children through their content, thereby exacerbating the mental health crisis among the youth \cite{lawsuit}. Video sharing websites combine both manual and automated content moderation approaches to eliminate the presence of published malicious content \cite{YouTubeContentMod,TikTokContentMod}. For this paper, we sampled the content currently available for pre-schoolers on YouTube Kids and identified a number of malicious videos that haven't been flagged by the existing content moderation system. Figures \ref{fig:ytkmachinegun} and \ref{fig:ytkpiano} show snapshots from two different problematic videos which include violence and disturbing music, respectively.   These examples highlight the need to develop robust and efficient tools for video content moderation.

Young children, particularly toddlers and pre-schoolers, are the demographic most susceptible to being adversely affected by video content created with malicious intent. Firstly, they are too naive to appreciate that certain audio-visual features are deliberately incorporated in videos in order to lure them to continue watching. Furthermore, continuous adult supervision is not pragmatic - in fact child's screen time serves as a babysitting tool, enabling parents to attend to other responsibilities or take a break, as reported by 32\% of parents \cite{babysitting_tool}. Numerous psychological studies discuss the adverse effects of exposure of inappropriate video content on a young child's mental and behavioral health.

\citeauthor{ahmed2023flairs} (\citeyear{ahmed2023flairs}) provide a catalog of features that characterize malicious video content including quick and repetitive movements, frightening or repulsive appearance, harmful or destructive actions, and inappropriate or indecent behavior. They also note that malicious audio may include high-volume music or noise, cries or shouts, sounds of explosions or gunshots, and the use of offensive language. Excessive exposure to these inappropriate videos poses a notable threat to the cognitive growth of preschoolers. 

Recent works on content moderation for children's videos have attempted to address the problem with multimodal approaches. \citeauthor{chuttur} (\citeyear{chuttur}) analyze text in the form of user comments and captions, along with image data. Their method is unable to capture the spatio-temporal context in videos as they use single images as inputs. \citeauthor{samba} (\citeyear{samba}) proposed a multimodal solution that incorporates subtitles and video metadata as input. As the subtitles do not contain information about music or sound effects, Samba may classify a video with malicious audio content as safe. \citeauthor{Papadamou} (\citeyear{Papadamou}) rely on video meta-data such as tags, title, statistics, in addition to video thumbnails. None of these models include audio inputs which is an essential modality for devising a robust video content moderation system. \citeauthor{tahir_audio} 
 (\citeyear{tahir_audio}) and \citeauthor{algh_audio} (\citeyear{algh_audio}) combined audio spectrograms with video features to classify the video content being either appropriate or inappropriate. However, the former focuses on detecting fake cartoon videos, whereas the latter only introduces a high-level idea without implementation.  

Our proposed work bridges the gap in previous works by adapting OpenAI's CLIP architecture \cite{vanilla_clip} for videos, and enabling it to learn prompts for both the vision and text branches. Furthermore, we incorporate a pre-trained audio encoder from AudioCLIP \cite{audio_clip} to handle the audio modality while adding a fully-trainable projection layer to improve performance on the downstream task of detecting malicious videos. We also introduce a multimodal dataset with both audio and video modalities that includes annotations specifically targeted for the problem of content moderation for children's videos.

Psychological research indicates that prolonged exposure to animated content featuring rapid movements can result in deterioration in performing everyday tasks \cite{fast_motion_psycho}. Likewise, the impact of loud noises on a child's brain development, particularly affecting reading, writing, and comprehension skills, has been established \cite{loud_noise_psycho}. The direct influence of violent cartoons on the behavior of preschoolers is evident, leading to increased levels of aggression and anxiety \cite{violent_cartoons_psycho}. Based on these psychological studies, in this paper we identify subtly harmful acoustic and visual features which makes the content alluring and addictive to a young child. These include a) brightly-hued fast moving objects with fast and loud music and b) split-screen animations with fast and loud music. A sample video snapshot for a) is shown in Figure \ref{fig:ytkpiano}.

This paper makes the following contributions towards the problem of content moderation of children's videos:
\begin{compactenum}
    \item Introduces a multimodal framework that includes the audio modality for more robust content moderation of children's cartoon videos. We perform ablations to demonstrate the enhanced performance of adding audio.
    
    \item Adapts CLIP to enable learning of prompts in both vision and text branches across multiple layers of encoders, along with AudioCLIP's pre-trained audio encoder.  We integrate a learnable projection layer in order to efficiently learn audio representations for the downstream content moderation task, while keeping the weights in the core models frozen.
    
    \item Presents a comprehensive multimodal dataset MMOB (\textbf{M}ultimodal \textbf{M}alicious \textbf{o}r \textbf{B}enign) which includes labeled cartoon video samples with annotations of vision and audio labels. Baseline results for MMOB in supervised and few-shot learning settings are also showcased. We have released the dataset to encourage further research in the area.
\end{compactenum}

\section{Background}\label{BG} 
\subsection{COPPA - Children's Online Privacy Protection Act} 
The Children's Online Privacy Protection Act (COPPA) \cite{COPPA1998}, enforced by the US Federal Trade Commission (FTC), establishes clear privacy and content guidelines for online content publishers targeting children aged 13 or younger. In 2019, additional provisions were incorporated into COPPA, particularly addressing YouTube, to guarantee the delivery of safe content for children. In January 2024, the Federal Trade Commission (FTC) released a proposal revising the regulations that enforce the Children's Online Privacy Protection Act (COPPA Rule). The proposed modifications aim to adapt to technological changes, and strengthen protections for children's personal information based on the FTC's review of public comments and enforcement experience \cite{FTC_2024}. Therefore, the problem of publishing age-appropriate content that we address in this paper, has been given high importance by FTC, which recommends avoiding content that could be harmful or inappropriate for children. 

\subsection{CLIP - Contrastive Language-Image Pre-Training}
CLIP \cite{vanilla_clip} is a very large multimodal model trained on a huge corpus of (image, text) pairs. Paired images and text are used for training both image and text encoders, jointly.  The user provides the CLIP model with the text prompt that will elicit the best image classification; this text may include ground truth class information. While learning the combined multimodal embedding space the model aims to maximize the cosine similarity scores of the image with the ground truth text while minimizing the cosine similarity between the embeddings of incorrect pairings. CLIP is capable of performing zero-shot classification and other downstream computer vision tasks efficiently using natural language supervision. In this work we modify CLIP for video input, and augment it to learn prompts in both the visual and textual branches of CLIP.

\begin{figure*}[htp]
    \centering
    \includegraphics[width=0.95\textwidth]{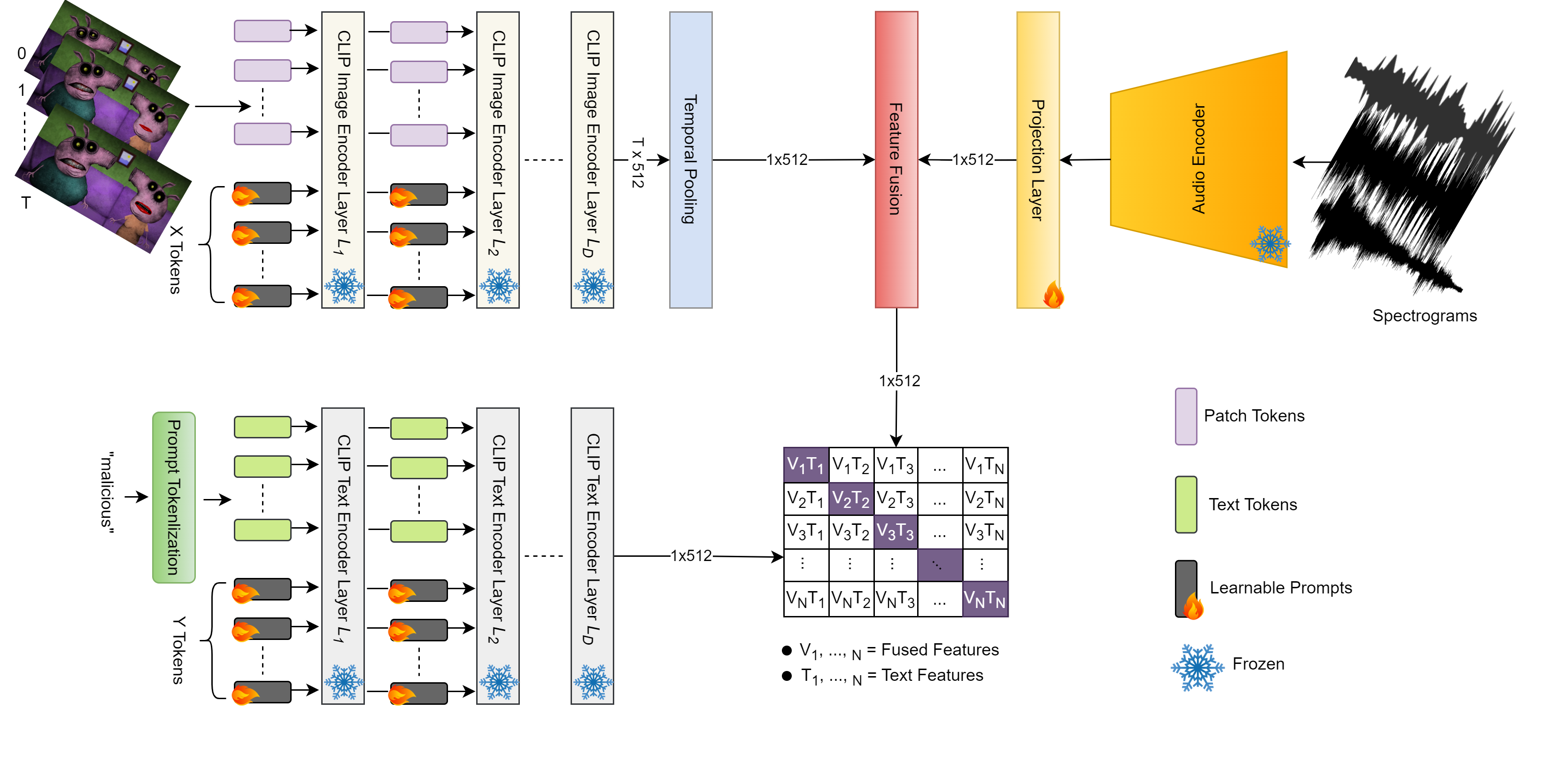}
    \caption{Our proposed architecture incorporates the audio modality by adding a pre-trained audio encoder. The inputs of the audio encoder are spectrograms which are visual representations of audio frequency signals. The trainable projection layer learns audio representations for the downstream content moderation task. Temporal pooling outputs a combined representation of \textit{T} input video frames, hence adapting Vanilla CLIP for video. These audio and visual representations are fused together within the Feature Fusion block. Vanilla CLIP's text and vision branches are adapted to include learnable prompts (tokens) through all layers. We keep all encoder layers of text, vision, and audio branches frozen. The input from the text branch is the class name, e.g. ``malicious'' as shown in the figure.}  
    \label{fig:model}
\end{figure*}

\section{Related Work}\label{RW}


\subsection{Audio Modality}
Audio is an oft neglected modality; however we believe that raw audio provides valuable contextual clues that are not included in the text transcription.  This is particularly true in children’s cartoons where sound effects are often used in lieu of words to convey meaning to young children with smaller vocabularies.  Popular cartoon sound effects include springs, tires screeching, slide whistles, chattering teeth, explosions, suction cups and rattles.   Previous work has looked explicitly at the problem of sound event classification~\cite{soundevent}, which could be potentially used to detect violent actions in children's cartoons.  A subcategory of this research focuses solely on environmental sound classification~\cite{esc1,esc2}, classifying the subset of sounds that people experience in the everyday environment such as road noise or music.   These background sounds can add important context to video and are robust to visual disturbances such as occlusion and illumination changes.

Audio representations have progressed significantly from the traditional signal processing representations such as FFTs or wavelets.  One popular representation is mel-frequency cepstral coefficients (MFCC) which are extracted from the discrete cosine transform of the log-mel spectrum~\cite{mfcc}.  In our work, we utilize the spectrogram which is a visual representation of the frequencies across time.   The spectrogram can serve as an input to convolutional neural networks or transformers.  The competition, Holistic Evaluation of Audio Representations~\cite{turian2022hear}, conducted a head to head evaluation of the benefits of different audio representations.  BYOL for Audio~\cite{niizumi_byol_2023} tackled the problem of learning audio representations that are suitable for a multiple downstream tasks.

Since there are few labeled audio datasets, the emergence of large multimodal datasets provides fertile opportunities for leveraging cross-modality information.  Several audio models have built on the success of the CLIP model at exploiting video and language.  For instance, Microsoft’s CLAP~\cite{elizalde_clap_2022} model learns a joint encoding for audio and text and can ingest text prompts to perform zero shot classification on audio data. 
    
\subsection{Prompt Learning}
Performing fine-tuning on large pre-trained models like CLIP \cite{vanilla_clip} leads to overfitting, which deteriorates the generalization of the model. Furthermore, it is also costly in terms of computation as all parameters of the model need to be updated. Prompt learning is an efficient approach where the pre-trained model parameters remain frozen while introducing learnable vectors called tokens. This helps in adapting the model to downstream tasks while while retaining the model's knowledge gained during pre-training. This approach is less computationally intensive  and requires less training time as compared to full fine-tuning a model. Prompt learning was introduced by researchers in the NLP area where the textual representations are similarly learned in the word embedding space \cite{nlp-prompt1,nlp-prompt2,nlp-prompt3,nlp-prompt4}. The approach was also adapted for vision tasks \cite{jia2022visual_16_deep,wang2022dualprompt_38,zhang2022neural_47} and for usage within vision-language models \cite{cocoop,coop,zhu2024promptaligned_51}.  \citeauthor{jia2022visual_16_deep} (\citeyear{jia2022visual_16_deep}) compare deep prompting in vision with shallow prompting and discuss the improvements in model performance due to the former contribution. Some works have discussed prompt learning in both text and vision branches \cite{maple,vifi,vita}. \citeauthor{maple} (\citeyear{maple}) also introduce a connection between vision and language prompts to ensure synergy while prompts are learnt.

\section{Methodology}\label{M}
Our proposed multimodal architecture amalgmates the visual, text and audio transformer encoders. The former two are adapted from Vanilla CLIP \cite{vanilla_clip} whereas the audio encoder is based on AudioCLIP \cite{audio_clip}. During training, all encoder layers of all modalities remain frozen, therefore retaining the model's pre-training knowledge while making it computationally efficient. To enhance the model's capability to be effective on the downstream task of video content moderation, prompt learning is enabled for the text and vision branches. We also introduce a fully-trainable projection layer at the end of the audio branch which learns the audio representations during training. Vanilla CLIP was tailored for the video input. We calculate the aggregate representations from both audio and vision branches.  The fused features along with the text features are then learned contrastively during training - maximizing the diagonal  (\textit{V$_{i}$},\textit{T$_{i}$}) of the cosine similarity matrix while minimizing other pairs of fused and text features. Maximizing the diagonal implies that the similarity score with the ground-truth class is maximized while others are minimized. The following sections discuss the important components of our proposed model.

\subsection{Adapting CLIP for Audio and Video}

\subsubsection{Pre-training Audio Encoder}
The pre-trained audio head which we include in our model is ESResNe(X)t \cite{esresnext} which is trained in four stages: 1) initialized with ImageNet weights \cite{image_net}, 2) fine-tuned on the AudioSet dataset \cite{audio_set} as a standalone model, 3) trained as part of the AudioCLIP model \cite{audio_clip} keeping text and vision branches frozen, and finally 4) fully trained with all three encoders of AudioCLIP fully trainable. 

\subsubsection{Training Projection Layer}\label{projlayerlabel}
In order to improve performance on the downstream task of classifying malicious and benign videos we introduce a fully-learnable projection layer while keeping the audio head frozen. Freezing the audio encoder layers not only helps in preventing catastrophic forgetting of the knowledge gained during pre-training but is also computationally efficient as the frozen parameters do not need to be updated during training. The projection layer dimensions are \textit{$1024 \times 512$}. ESResNe(X)t uses the ResNet50 backbone \cite{resnet50} which has an embedding dimension of \textit{1024}.  To make the output dimension compatible with CLIP's vision backbone we choose \textit{512} as the projection layer's output dimension size.

\subsubsection{Adapting CLIP for Video}
CLIP is jointly pre-trained on images and text; in order to train on video input we perform Temporal Pooling. The result of Temporal Pooling is a combined representation of all \textit{T} input frames of a video; the temporal information is accumulated by averaging frame-level features. Hence, the representation of visual features of a video with dimension \textit{$T \times 512$} transforms to \textit{$1 \times 512$}. 
Here, the dimension 512 denotes the embedding size of the vision transformer base model ViT-B/16. 

\subsubsection{Feature Fusion}
The output of the audio and vision encoders are individual representations of the acoustic and visual features respectively. Mixing both types of feature representations provides the model with additional context about the scene. During feature fusion we construct a joint representation by adding the audio and visual feature representations. The input (visual and audio) and the fused embedding dimensions are \textit{$1 \times 512$} .

\subsection{Learning Prompt Representations}\label{promptlearninglabel}

To reduce the training time and computation cost of fine-tuning CLIP for the downstream task of video content moderation we disable the learning of the pre-trained text and video encoder parameters and employ learnable tokens on both branches. Figure \ref{fig:model} illustrates the proposed model. Prompting in both branches helps prompt tokens learn the context more effectively as the prompts adapt textual, in addition to vision representations jointly. Furthermore, we perform the learning of prompts in multiple layers, referred to as deep prompting \cite{maple}. Learning deep prompts especially helps in boosting performance for low data regimes. Since our benchmark dataset, MMOB, is not very large, deep prompting has the potential to improve performance as compared to shallow prompting where only prompting is performed in one or few layers. The dataset is available at \url{https://github.com/syedhammadahmed/mmob}.

A series of experiments was conducted to explore how activation or deactivation of prompt learning in the vision branch of the adapted CLIP model affects overall accuracy. We also investigate whether results improve if we increase 1) the depth of prompt token training, and 2) the number of tokens. The results show that having prompt learning enabled in both text and vision branches of our adapted CLIP model gives the best accuracy. Similarly, increasing the number of prompt tokens in a layer improves the classification accuracy. A similar trend was observed when we increase the learning depth in terms of layers. The detailed discussion on these ablations is included in the Experiments section of this paper.

\begingroup
\renewcommand{\arraystretch}{1.6}
\begin{table}[h]
\centering

\begin{tabularx}{0.4\textwidth} { 
   >{\centering\arraybackslash}X 
  | >{\centering\arraybackslash}X 
  | >{\centering\arraybackslash}X}
 \multicolumn{3}{c}{} \\
 \hline
 \textbf{Malicious} & \textbf{Benign} & \textbf{Total}\\ 
 \hline
 \hline
 305 & 830 & 1135\\
\end{tabularx}
\caption{Distribution of malicious and benign videos in the MMOB dataset.}
\label{table:dataset}
\end{table}
\endgroup

\section{Experiments}\label{BE}
\subsection{The Multimodal Malicious or Benign Dataset}
We curate the \textbf{M}ultimodal \textbf{M}alicious \textbf{o}r \textbf{B}enign (MMOB) dataset by selecting the samples containing malicious videos with malicious audio tracks, and benign videos with benign audio tracks. The samples have been adapted from the MOB dataset \cite{ahmed2023flairs}. MOB is a cartoon video dataset including visual annotations only. We extract the audio from the videos and generate the test-train splits for this new multimodal dataset. 
Table \ref{table:dataset} shows the class-wise sample counts.

\subsection{Training and Evaluation} 
We evaluate the model with the MMOB dataset in both supervised and few-shot settings. For the supervised learning setting, we use the full training split of our dataset while for the few shot learning settings, a subset of $k_i = {0, 1, 2, 4, 8, 16}$ videos is randomly sampled from the training split. For all experiments, we use ViT-B/16 as our base model which is pre-trained using CLIP. 

For the default setup, we employ 12 layers of prompt learning for both the CLIP Text encoder and CLIP Video encoder. Furthermore, each encoder utilizes prompt learning with 12 tokens. It's worth noting that, in our case, the projection layer at the end of the audio head is learned during the training process.


During the training, we take 16 frames for each video along with the associated sound spectograms and use the class label as text. We also pass on learnable prompts to both video encoder and text encoder where $X$ represents the number of tokens used by the video encoder while $Y$ represents the number of tokens for text encoder. For most of the experiments, the default value of 12 tokens for both video encoder and text encoder is used. However, to analyze the impact of these learnable prompts, we evaluate the effect of decreasing the number of tokens.  The learning rate is set to $8\times10^{-5}$ in all experiments. Lastly, the depth of the encoders is represented by $D$ as depicted in Figure \ref{fig:model}.

\subsection{Results and Discussion}

In this section, we present our empirical analysis using the pre-trained Vanilla CLIP model, where the base model ViT-B/16 is used, unless explicitly mentioned otherwise. Throughout our experiments, we leverage all modalities for both base models. Furthermore, we maintain the default settings for the number of tokens and the depth of text prompt layers i.e. 10 and 12, respectively. Notably, ViT-B/16 demonstrated superior performance compared to ViT-B/32, achieving an improvement of over 5.5\%. The reason ViT-B/16 may exhibit better performance than ViT-B/32 can be attributed to the finer details captured by the smaller patch size. We fine-tune Vanilla CLIP with both base models for 20 epochs and report the accuracy in Table \ref{table:main_res}.

\begingroup
\renewcommand{\arraystretch}{1.6}
\begin{table}[h]
\centering

\begin{tabularx}{0.47\textwidth} { 
   >{\centering\arraybackslash}X 
  | >{\centering\arraybackslash}X 
  | >{\centering\arraybackslash}X
  | >{\centering\arraybackslash}X 
  | >{\centering\arraybackslash}X 
  | >{\centering\arraybackslash}X
  | >{\centering\arraybackslash}X}
 \multicolumn{3}{c |}{\textbf{Modalities}} & \multicolumn{3}{c |}{\textbf{Learnable}} & \multirow{2}{5em}{\textbf{Acc}}\\ 
 \cline{1-6} 
 \textbf{Text} & \textbf{Video} & \textbf{Audio} & \textbf{Text} & \textbf{Video} & \textbf{Audio} & \\
 \hline
 \hline
 \textbf{\checkmark} & \textbf{\checkmark} & \textbf{\checkmark} & \textbf{\checkmark} & \textbf{\checkmark} & \textbf{\checkmark} & \textbf{81.49}\\
 
 \textbf{\checkmark} & \textbf{\checkmark} & \textbf{\ding{55}} & \textbf{\checkmark} & \textbf{\checkmark} & \textbf{\ding{55}} & 78.41 \\ 
 \textbf{\checkmark} & \textbf{\checkmark} & \textbf{\ding{55}} & \textbf{\checkmark} & \textbf{\ding{55}} & \textbf{\ding{55}} & 76.65 \\ 
 \textbf{\checkmark} & \textbf{\checkmark} & \textbf{\checkmark} & \textbf{\checkmark} & \textbf{\ding{55}} & \textbf{\checkmark} & 78.21 \\ 
\end{tabularx}
\caption{Adding modality and learning prompts (text/video) along with the projection layer (audio) improves the overall accuracy of the model.}
\label{table:modalities}
\end{table}
\endgroup

\begingroup
\renewcommand{\arraystretch}{1.6}
\begin{table*}[htp]
\centering

\begin{tabularx}{\textwidth} { 
  >{\raggedright\arraybackslash}X 
  | >{\centering\arraybackslash}X 
  | >{\centering\arraybackslash}X 
  | >{\centering\arraybackslash}X
  | >{\centering\arraybackslash}X 
  | >{\centering\arraybackslash}X 
  | >{\centering\arraybackslash}X
  | >{\centering\arraybackslash}X 
  | >{\centering\arraybackslash}X 
  | >{\centering\arraybackslash}X 
  | >{\centering\arraybackslash}X 
  | >{\centering\arraybackslash}X }
 \multirow{2}{5em}{\textbf{Base \quad Model}} & \multicolumn{3}{c |}{\textbf{Modalities}} & \multicolumn{3}{c |}{\textbf{Learnable Prompts}} & \multicolumn{2}{c |}{\textbf{No. of Tokens}} & \multicolumn{2}{c |}{\textbf{Prompt Depth}} & \multirow{2}{5em}{\textbf{Accuracy}}\\ 
 \cline{2-11} 
 & \textbf{Text} & \textbf{Video} & \textbf{Audio} & \textbf{Text} & \textbf{Video} & \textbf{Audio} & \textbf{Text} & \textbf{Video} & \textbf{Text} & \textbf{Video} & \\
 \hline
 \hline
 ViT-B/16 & \textbf{\checkmark} & \textbf{\checkmark} & \textbf{\checkmark} & \textbf{\checkmark} & \textbf{\checkmark} & \textbf{\checkmark} & 12 & 12 & 12 & 12 & \textbf{81.49} \\
 ViT-B/32 & \textbf{\checkmark} & \textbf{\checkmark} & \textbf{\checkmark} & \textbf{\checkmark} & \textbf{\checkmark} & \textbf{\checkmark} & 12 & 12 & 12 & 12 & 77.09
 
\end{tabularx}
\caption{Results of fine-tuning CLIP on our MMOB dataset with different base models and learnable prompts. ViT-B/16 achieves better accuracy in comparison to the other model. Both models have been fine-tuned for 20 epochs.}
\label{table:main_res}
\end{table*}
\endgroup

\subsubsection{Impact of Adding Audio Modality and Learnable Prompts}

We evaluate the improvement derived from adding the audio modality, deep learnable prompts on both vision and text branches, and enabling the learning of the projection layer deployed at the end of the audio encoder. Our approach involves systematically eliminating modalities and learnable modules, one at a time in each experiment. To better understand the impact of these configuration changes we maintain a consistent number of prompts and depth of prompt layers. The ensuing analysis of the model's performance is presented in Table \ref{table:modalities}. As anticipated, the inclusion of additional modalities and prompt learning contributes positively to the overall accuracy. The baseline accuracy of 81.49 \% in the supervised setting was observed in the configuration where we include 1) audio modality with video, 2) prompting in language and vision branches, and 3) learning of the audio projection layer.    

\subsubsection{Impact of Token Length}

To delve deeper into the model's performance, we systematically diminish the number of tokens in our investigation. In this set of experiments, we maintain the utilization of all three modalities, each equipped with learnable prompt layers. The sole variation lies in the adjustment of token quantity. Table \ref{table:tokens} provides a comprehensive summary of the results derived from these experiments. Using a greater number of tokens positively contributes to the learning of the model. 

\begingroup
\renewcommand{\arraystretch}{1.6}
\begin{table}[h]
\centering

\begin{tabularx}{0.4\textwidth} { 
   >{\centering\arraybackslash}X 
  | >{\centering\arraybackslash}X 
  | >{\centering\arraybackslash}X}
 \multicolumn{2}{c |}{\textbf{No. of Tokens}} & \multirow{2}{5em}{\textbf{Accuracy}}\\ 
 \cline{1-2} 
 \textbf{Text} & \textbf{Video} &  \\
 \hline
 \hline
 10 & 10 & \textbf{81.49}\\ 
 10 & 8 & 80.17\\ 
 10 & 6 & 79.29\\ 
 10 & 4 & 77.97\\ 
\end{tabularx}
\caption{Increasing the number of learnable prompt tokens improves performance. }
\label{table:tokens}
\end{table}
\endgroup

\subsubsection{Impact of Prompt Learning Depth}
The role of prompt learning is pivotal in recent vision-language models. In this series of experiments, we specifically explore the influence of prompt learning depth. Once more, we employ all three modalities, maintaining a consistent token count of 10 for both text and video heads. The outcomes of these experiments are detailed in Table \ref{table:prompt}. A discernible pattern emerges, indicating that as the depth of prompt learning increases, there is a corresponding enhancement in the model's accuracy.

\begingroup
\renewcommand{\arraystretch}{1.6}
\begin{table}[h]
\centering

\begin{tabularx}{0.4\textwidth} { 
   >{\centering\arraybackslash}X 
  | >{\centering\arraybackslash}X 
  | >{\centering\arraybackslash}X}
 \multicolumn{2}{c |}{\textbf{Depth of Prompt Learning}} & \multirow{2}{5em}{\textbf{Accuracy}}\\ 
 \cline{1-2} 
 \textbf{Text} & \textbf{Video} &  \\
 \hline
 \hline
 12 & 12 & \textbf{81.49}\\ 
 8 & 8 & 80.61\\ 
 4 & 4 & 77.97\\ 
 2 & 2 & 75.77\\ 
\end{tabularx}
\caption{Multiple depth levels for prompt learning also enhance model's performance.}
\label{table:prompt}
\end{table}
\endgroup

\subsubsection{Few-shot Setting}

Given that Vanilla CLIP is a pretrained network, it lends itself well to few-shot learning. In these experiments, we introduce the few-shot learning parameter, denoted as $k$, with values specified as $k_i = \{0, 1, 2, 4, 8, 16\}$. This means that for each class, we utilize $k_i$ samples, while keeping the other parameters set to their original values. The samples are drawn uniformly from each class without replacement.

\begingroup
\renewcommand{\arraystretch}{1.6}
\begin{table}[h]
\centering

\begin{tabularx}{0.4\textwidth} { 
   >{\centering\arraybackslash}X 
  | >{\centering\arraybackslash}X}
 \textbf{$k_i$} & \textbf{Accuracy}\\ 
 \hline
 \hline
 0 & 59.47\\
 1 & 63.01\\
 2 & \textbf{64.75}\\ 
 4 & 63.87\\ 
 8 & 61.67\\ 
 16 & 54.18\\ 
\end{tabularx}
\caption{Few shot learning for various $k$ samples of each class.}
\label{table:kshot}
\end{table}
\endgroup

In few shot learning experiments, we see an interesting trend. For $k = 2$, the model achieves the highest accuracy while as the value of $k$ increases, the accuracy drops. This trend is shown in Table \ref{table:kshot}.  The choice of $k$ affects the intrinsic dimensionality of the learned embedding space, hence increasing $k$ does not result in monotonic performance improvements~\cite{fewshot}.

\section{Conclusion and Future Work}
This paper highlights the importance of leveraging the audio modality for the problem of content moderation of children's videos. We introduce a new multimodal architecture that adds a pre-trained audio encoder with a learnable projection layer to the adapted CLIP model. Our adapted model averages video frames to create a visual representation and learns prompts on both branches of Vanilla CLIP (text and vision). This contribution enhances the overall training and performance of video content moderation.  We also created a multimodal dataset, MMOB, for this task that includes audio annotations.  In future work, we intend to extend our work to other types of online video content including YouTube shorts, Facebook reels, and TikTok videos. These short duration videos are popular with viewers but may create negative impacts such as reducing attention span. Lastly, we plan to work on the problem of moderating the advertisements that accompany children's videos.  From personal experience browsing these platforms, we observed that although a video may be appropriate for viewing by young children the advertisements occasionally contain unsuitable content.
\newpage
\bibliographystyle{flairs} 
\bibliography{references.bib}

\end{document}